\documentclass[runningheads]{llncs}
\usepackage[T1]{fontenc}
\usepackage{graphicx}

\usepackage{amsmath,amssymb} 
\usepackage{makecell}
\usepackage{verbatim}
\usepackage{hyperref}
\hypersetup{hidelinks,
	colorlinks=true,
	allcolors=black,
	pdfstartview=Fit,
	breaklinks=true}

\begin{document}
\title{D$^{\bf{3}}$: Duplicate Detection Decontaminator \\for Multi-Athlete Tracking in Sports Videos}
\titlerunning{D$^{\bf{3}}$: Duplicate Detection Decontaminator}
%
\author{Rui He\inst{1}\orcidID{0000-0002-5138-1654} \and
Zehua Fu\inst{1,2}\orcidID{0000-0002-3639-4406} \and\\
Qingjie Liu\inst{1,2}\thanks{Corresponding Author}\orcidID{0000-0002-5181-6451} \and 
Yunhong Wang\inst{1}\orcidID{0000-0001-8001-2703} \and
Xunxun Chen\inst{3}\orcidID{0000-0002-9481-4819}}
\authorrunning{R. He et al.}
%
\institute{Laboratory of Intelligent Recognition and Image Processing (IRIP Lab),\\Beihang University (BUAA), Xueyuan Road No.37, Haidian District, Beijing, China \\
\and
Hangzhou Innovation Institute, Beihang University, Hangzhou, China \\
\and
National Computer Network Emergency Response Technical Team/Coordination Center of China (CNCERT or CNCERT/CC), Beijing, China \\
\email{\{heruihr,zehua\_fu,qingjie.liu,yhwang\}@buaa.edu.cn}, 
\email{cxx@cert.org.cn}}
\maketitle              
\begin{abstract}
Tracking multiple athletes in sports videos is a very challenging Multi-Object Tracking (MOT) task, since athletes often have the same appearance and are intimately covered with each other, making a common occlusion problem becomes an abhorrent duplicate detection. In this paper, the duplicate detection is newly and precisely defined as occlusion misreporting on the same athlete by multiple detection boxes in one frame. To address this problem, we meticulously design a novel transformer-based Duplicate Detection Decontaminator (D$^3$) for training, and a specific algorithm Rally-Hungarian (RH)  for matching. Once duplicate detection occurs, D$^3$ immediately modifies the procedure by generating enhanced boxes losses. RH, triggered by the team sports substitution rules, is exceedingly suitable for sports videos. Moreover, 
to complement the tracking dataset that without shot changes, we release a new dataset based on sports video named RallyTrack. Extensive experiments on RallyTrack show that combining D$^3$ and RH can dramatically improve the tracking performance with 9.2 in MOTA and 4.5 in HOTA. Meanwhile, experiments on MOT-series and DanceTrack discover that D$^3$ can accelerate convergence during training, especially save up to 80 percent of the original training time on MOT17. Finally, our model, which is trained only with volleyball videos, can be applied directly to basketball and soccer videos for MAT, which shows priority of our method. Our dataset is available at \href{https://github.com/heruihr/rallytrack}{https://github.com/heruihr/rallytrack}.

\end{abstract}

\section{Introduction}

Sports video analysis possesses wide application prospects and is currently receiving plenty of attention from academia and industry. Scene understanding in sports video can be utilized for data statistics\cite{GiancolaADG18}, highlight extraction\cite{NiuGT12}, tactics analysis\cite{KongZRLH21}. Multi-Athlete Tracking (MAT)\cite{KongHW20} is a basic task in sports video-based scene understanding, which occupies a pivotal position.

\begin{figure}[t]
\centering
\begin{tabular}{cc}
\includegraphics[width=7.15cm]{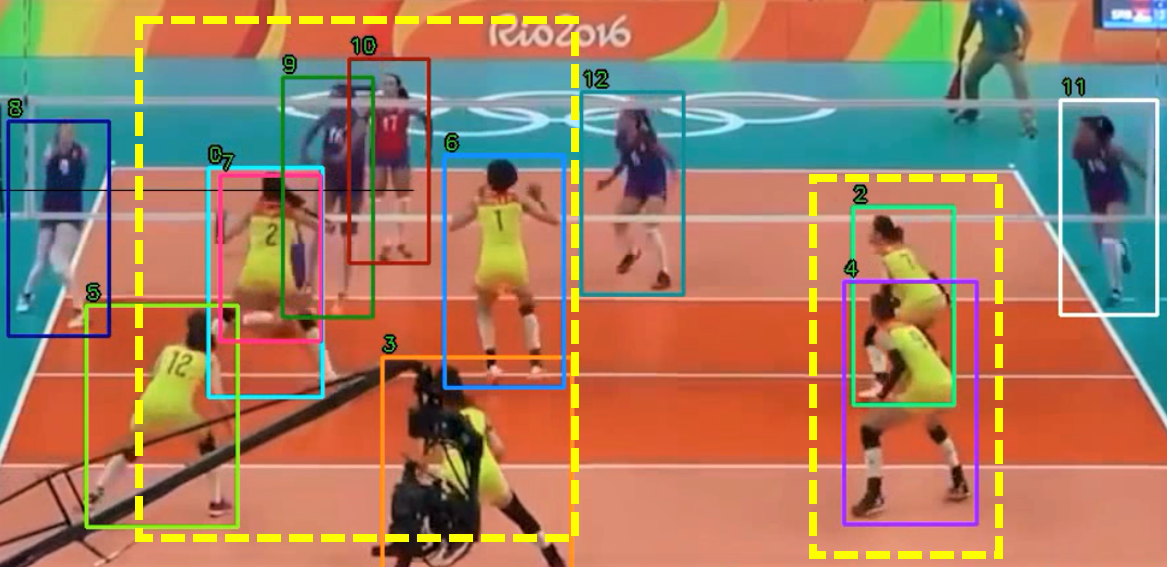}&
\includegraphics[width=3.7cm]{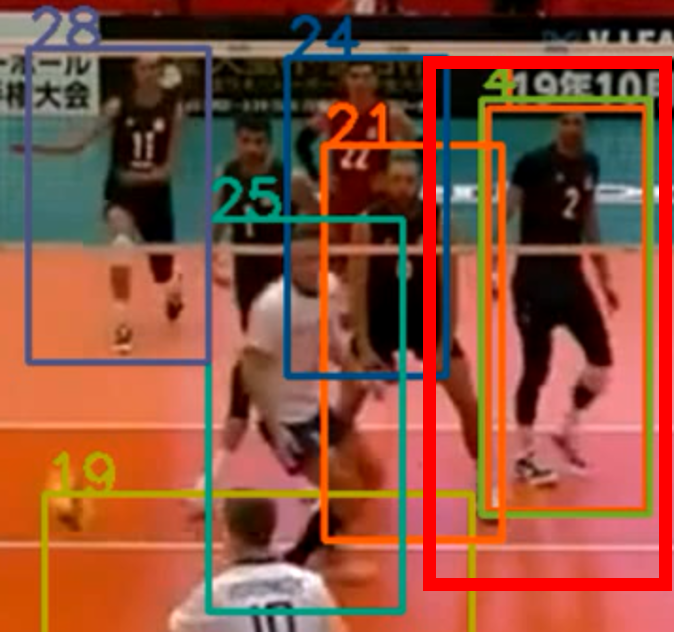}\\
(a)&(b)
\end{tabular}
\caption{A labeled sample in RallyTrack and the duplicate detection problem occurred in a TransTrack model: (a) a labeled sample with heavy occlusions, label\_7 is totally covered by label\_0 in the left dash box; (b) red box shows duplicate detection, the same individual is detected by two queries with two IDs.}
\label{fig:intro}
\end{figure}

Unlike general Multi-Object Tracking (MOT)\cite{MilanL0RS16,abs-2003-09003}, in MAT, different athletes generally share a high similarity in appearance and they often have a diversity of action changes and abrupt movements. The former difficulty, from our observation, turns a common occlusion problem in MOT to duplicate detection in sports video, which is defined in this paper as occlusion misreporting on the same object by multiple predictions in the same frame. The latter one leads to objects detetion missing, which often accompany with duplicate detection. In contrast to general person-based MOT, Figure \ref{fig:intro}(a) displays the difficulties with two yellow dash boxes. These two main difficulties make MAT a challenging task.

Then Figure \ref{fig:intro}(b) illustrates duplicate detection with a red box, which may be caused by two possibilities. One is that all athletes are detected, and an athlete being repeatedly detected is treated as occluding an additional invisible athlete. The other is that not all athletes are detected, some detections are missing while someone is repeatedly detected.  
To address this issue, we design a Duplicate Detection Decontaminator (D$^3$) , which can keep watch on the training procedure. Once a duplicate detection occurs, D$^3$ can generate additionally enhanced self-GIoU\cite{RezatofighiTGS019} losses during training. Then the losses will be gradually backpropagated to force the duplicate detecting boxes to keep away from each other. When duplicate detection disappears, D$^3$ would not produce loss anymore. We also offer a specific matching algorithm called Rally-Hungarian (RH) algorithm for MAT, which is triggered by the substitution rule in team sports like volleyball. What is more, to make up for the lack of shot change, a new dataset namely RallyTrack is annotated, which is based on a scene of sports video,  and Figure \ref{fig:intro}(a) is a labeled sample.

Intensive experimental results on RallyTrack demonstrate the efficiency of D$^3$ and RH. During our experiments, we discover that duplicate detection is not only a prominent problem in MAT, but also an unnoticed barrier often hidden in MOT, which makes model converge slowly. Experimental results on MOT17\cite{MilanL0RS16} show D$^3$  can save up to 80 percent of original training time. More experiments on  MOT16\cite{Leal-TaixeMRRS15}, MOT20\cite{abs-2003-09003}, and DanceTrack\cite{abs-2111-14690} also confirm the priority of our method. 

The main contributions of this study are as follows.
(i) We design a Duplicate Detection Decontaminator (D$^3$) which supervises the training procedure to optimize detection and tracking boxes.
(ii) We design a matching algorithm called Rally-Hungarian (RH)  for MAT to further improve tracking result.
(iii) We annotate a new dataset named RallyTrack, which is based on scenes of sports videos, to make up for the lack of videos without shot change.
(iv) We perform extensive experiments to demonstrate and verify that the proposed method improves the tracking performance on MAT with a total enhancement of 9.2 for MOTA and 4.5 for HOTA, and D$^3$ can accelerate training convergence on MOT.

\section{Related Work}

\subsection{Multiple Object Tracking Datasets}

{\bf{Human-based Datasets.}} Concentrating on variant scenarios, a large number of multiple object tracking datasets have been collected, and  human tracking datasets accounted for a big proportion. One of the earliest datasets in this area is PETS  \cite{EllisF10} . Then MOT15 \cite{Leal-TaixeMRRS15} and the recent  MOT17 \cite{MilanL0RS16} , MOT20  \cite{abs-2003-09003} datasets become more and more popular in this community.  MOT datasets mainly contain handful of pedestrian videos, which are limited in regular movements of objects  and distinguishable appearance. As a consequence of that,  multiple object tracking could be easily achieved with association by pure appearance matching \cite{PangQLCLDY21} . More recently DanceTrack  \cite{abs-2111-14690} is proposed as a large-scale dataset for multi-human tracking, where humans have similar appearance, diverse motion and extreme articulation, which is expected to make research rely less on visual discrimination and depend more on motion analysis. 
However, the background in DanceTrack is usually identifiable to the foreground so that detecting is easy and tracking is hard. Collected from real and noisy match videos, our dataset is both challenging in detecting and tracking. 

{\bf{Diverse Datasets.}} Besides,  WILDTRACK \cite{ChavdarovaBBMJB18} , Youtube-VIS \cite{abs-1809-03327}, and MOTS  \cite{VoigtlaenderKOL19} are proposed for diverse objectives. With the development of autonomous driving,  KITTI \cite{GeigerLU12}, one of the earliest large-scale multiple object tracking datasets for driving, is interested in vehicles and pedestrians. Then larger scale autonomous driving datasets BDD100K\cite{YuCWXCLMD20} and Waymo\cite{SunKDCPTGZCCVHN20} are published. Limiting by lanes and traffic rules, the motion patterns of objects in these datasets are even more regular than moving people. Meanwhile, some datasets broaden their horizon on more diverse object categories. Trajectory annotations for 30 object categories in over 1000 videos are released by the ImageNet-Vid \cite{DengDSLL009} benchmark and TAO \cite{DaveKTSR20} annotates 833 object categories to study object tracking on long-tailed distribution.

\subsection{Object Detection in MOT}

{\bf{Tracking by detection.}} Object detection \cite{LinGGHD17,abs-1804-02767,RenHGS15} develops so vigorously that a lot of methods would like utilizing powerful detectors to pursue higher tracking performance. RetinaTrack \cite{LuRVH20} and ChainedTracker \cite{PengWWWWTWLHF20} apply the one-stage object detector RetinaNet \cite{LinGGHD17} for tracking. For its simplicity and efficiency, CenterNet  \cite{abs-1904-07850} becomes a popular detector adopted by CenterTrack \cite{ZhouKK20} and FairMOT \cite{ZhangWWZL21} . The YOLO series detectors  \cite{abs-1804-02767} are also put to use by TransMOT \cite{abs-2104-00194}  due to its excellent balance of accuracy and speed. The common ground of these methods is that the detection boxes are usually directly employed on a single image for tracking. However, as is pointed out by \cite{TangWWLZW20},  when occlusion or motion blur happens, the increasing number of missing detections and very low scoring detections would influence the quality of object linking. Therefore, the information of the previous frames are usually leveraged to enhance the video detection performance. Versatile Affinity Network (VAN) \cite{LeeK020}  is proposed to handle incomplete detection issues, affinity computation between targetand candidates, and decision of tracking termination.
\cite{HoKPKK20}  presents an approach, which is completely supervision-free in the sense that no human annotations of person IDs are employed,  injecting spatio-temporally derived information into convolutional AutoEncoder in order to produce a suitable data embedding space for multipleobject tracking.

{\bf{Joint-detection-and-tracking}}. Achieving detection and tracking simultaneously in a single stage is the destination of  the joint-detection-and-tracking pipeline. Aiming to enhance detection results, some early methods \cite{abs-2104-00194} utilize single object tracking (SOT) \cite{BertinettoVHVT16} or Kalman filter \cite{1960A}to predict the location of the tracklets in the following frame, and fuse the predicted boxes with the detection boxes.  Then by combining the detection boxes in the current frame and tracks in previous frames, Integrated-Detection \cite{abs-1811-11167} boosts the detection performance. Recently, Tracktor \cite{BergmannML19} directly uses the previous frame tracking boxes as region proposals, and then applies the bounding box regression to provide tracking boxes on the current step, thus eliminating the box association procedure. From a shared backbone, JDE \cite{WangZLLW20} and FairMOT \cite{ZhangWWZL21} learn the object detection task and appearance embedding task in the meantime. Different from CenterTrack \cite{ZhouKK20} localizing objects by tracking-conditioned detection and predicting their offsets to the previous frame, ChainedTracker \cite{PengWWWWTWLHF20} chains paired bounding boxes estimated from overlapping nodes, in which each node covers two adjacent frames. More recently, for its strong ability to propagate boxes between frames, transformer-based \cite{VaswaniSPUJGKP17} detectors like DETR \cite{CarionMSUKZ20,ZhuSLLWD21} are adopted by several methods, such as TransTrack \cite{abs-2012-15460},TrackFormer \cite{abs-2101-02702} , and MOTR \cite{abs-2105-03247} . Our method also follows this stucture to utilize the similarity with tracklets to strength the reliability of detection boxes. 

\subsection{Data Association}

{\bf{Tracking by matching appearance.}}
Appearance similarity is useful in the long-range matching, and serves as linchpin in many multi-object tracking methods.  DeepSORT \cite{WojkeBP17} adopts a stand-alone Re-ID model to extract appearance features from the detection boxes. POI \cite{YuLLLSY16} achieves excellent tracking performance depended on the high quality detection and deep learning-based appearance feature.  Recently, because of their simplicity and efficiency, joint detection and Re-ID models, such as RetinaTrack \cite{LuRVH20},  QuasiDense (QDTrack) \cite{PangQLCLDY21}, JDE \cite{WangZLLW20}, FairMOT  \cite{ZhangWWZL21}, becomes more and more prevalent. Designing a pairwise training paradigm and dense localization for object detection, QuasiDense (QDTrack) \cite{PangQLCLDY21}  compares highly sensitive appearance of objects across frames for matching.  For better appearance representation, JDE \cite{WangZLLW20} and FairMOT \cite{ZhangWWZL21} learn object localization and appearance embedding with a shared backbone.

{\bf{Tracking with motion analysis.}}
Object tracking usually regards the displacement of objects-of-interest as dominate cues. A line of researches has thus been inspired by a natural and intuitive idea that track objects by estimating their motion.  SORT \cite{BewleyGORU16} first adopts Kalman filter \cite{1960A} to predict the location of the tracklets in the new frame, and then by Hungarian algorithm \cite{1955The} computes the IoU between the detection boxes and the predicted boxes as the similarity for tracking. STRN \cite{Xu0ZH19} presents a similarity learning framework between tracks and objects, which encodes various Spatial-Temporal relations.  
Tracking by associating almost every detection box instead of only the high score ones, for the low score detection boxes, ByteTrack \cite{abs-2110-06864} utilizes their similarities with tracklets to recover true objects and filter out the background detections. Recently attention mechanism \cite{VaswaniSPUJGKP17} can directly propagate boxes between frames and perform association implicitly. TransTrack \cite{abs-2012-15460} is designed to learn object motions and achieves  robust results in cases of large camera motion or low frame rate.  TrackFormer \cite{abs-2101-02702}  and MOTR \cite{abs-2105-03247} propose track queries to find the location of the tracked objects in the following frames. 

\section{Duplicate Detection Decontaminator and Rally-Hungarian Algorithm}
\label{sec:DDDRH}

\begin{figure}
\centering
\includegraphics[width=10.55cm]{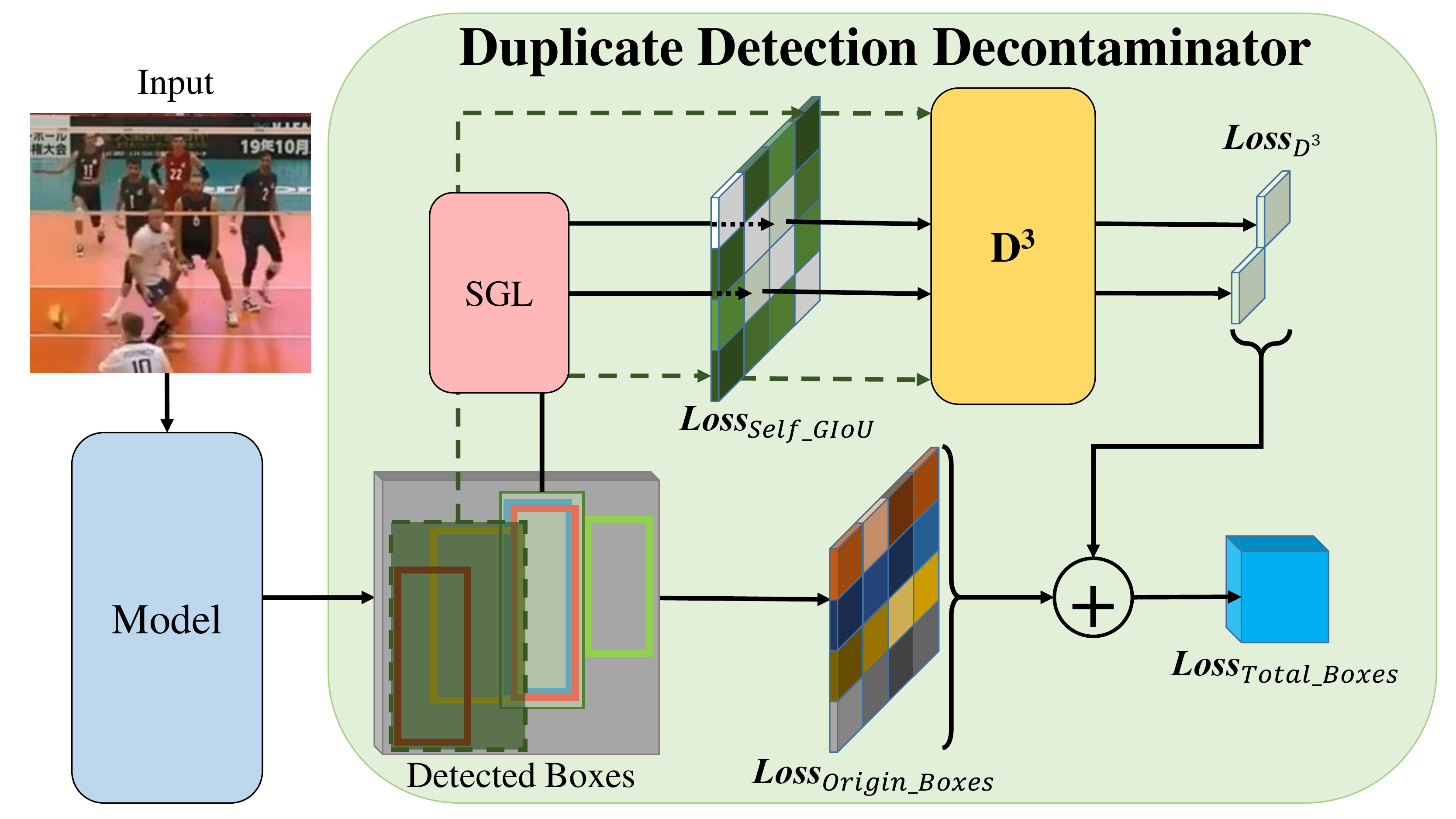}
\caption{While in the training stage, a Self-GIoU Loss matrix is constructed according to the detected boxes of the input frame $t$. Then D$^3$ will set a Lower Bound (LB) to the check self-GIoU loss. Values lower than LB in the matrix regarded as duplicate detection will be output and added to the origin boxes loss as a total boxes loss to be backpropagated. If there is no duplicate detection, D$^3$ will do nothing.}
\label{fig:DDD}
\end{figure}

In this section, the working mechanism of Duplicate Detection Decontaminator (D$^3$) and Rally-Hungarian (RH) matching algorithm will be introduced in detail respectively. Our methods are based on transformer, an encoder-decoder framework. The encoder takes the composed feature maps of two consecutive frames as input and generates keys for decoder. Working in a way of joint-detection-and-tracking, two parallel decoders, detection decoder and tracking decoder, are employed. D$^3$ is applied in the two decoders during training stage to optimize quality of detection boxes and tracking boxes. Then RH, a box IoU matching method, is utilized to obtain the final tracking result.

\subsection{Duplicate Detection Decontaminator}
\label{sec:DDD}
In a transformer-based tracking model, objects in an image are detected by harnessing learned object queries, which is a set of learnable parameters trained together with all other parameters in the network. While training a model, duplicate detection appears and D$^3$ will unroll its power then.

At training stage as shown in Figure \ref{fig:DDD}, denote $B=\{{\bf{b}}_{i} |i=1,\cdots,N\}$ as the boxes set of individuals in the middle output of input frame $t$, where ${\bf{b}}_{i}=(x_{i}^{1},y_{i}^{1},x_{i}^{2},y_{i}^{2})$ indicates the top-left corner $(x_{i}^{1},y_{i}^{1})$ and bottom-right corner $(x_{i}^{2},y_{i}^{2})$ of $i$th individual. Then applying the concept of Generalized Intersection over Union (GIoU)  \cite{RezatofighiTGS019}, we get Self-GIoU from $B$, where the element in $GIoU(B,B)$ is formulated as follows:
\begin{align}
\label{giou}
GIoU({\bf{b}}_i,{\bf{b}}_j) = \frac{|{\bf{b}}_i \cap {\bf{b}}_j|}{|{\bf{b}}_i \cup {\bf{b}}_j|} - \frac{|C \setminus ({\bf{b}}_i \cup {\bf{b}}_j)|}{|C|}
=IoU({\bf{b}}_i,{\bf{b}}_j)  - \frac{|C \setminus ({\bf{b}}_i \cup {\bf{b}}_j)|}{|C|}
\end{align}
where $C$ is the smallest convex hull that encloses both ${\bf{b}}_i$ and ${\bf{b}}_j$, IoU means Intersection over Union. Then we construct Self-GIoU Loss (SGL) matrix ${\bf{B}}$ as following:
\begin{align}
{\bf{B}} = SGL(B) = 1 - GIoU(B,B)
\end{align}
{\bf{B}} is a symmetric matrix in which the elements on the diagonal of the matrix are all 0, as painted white in Figure \ref{fig:DDD}. Then D$^3$ will set a Lower Bound (LB) to check {\bf{B}}. Once a value of the element in {\bf{B}} is lower than the LB, which means duplicate detection happens, D$^3$ will output the value and add it to the original boxes loss as a total boxes loss to be backpropagated. If there is no duplicate detection, D$^3$ will do nothing. The mechanism of D$^3$ is as follows:
\begin{align}
Loss_{D^3} = D^3({\bf{B}}) = \frac{1}{2}\sum_{i=1}^{N}\sum_{j=1}^{N}{\bf{B}}_{ij}, \quad {\bf{B}}_{ij} < LB \\
Loss_{Total\_Boxes} = Loss_{Origin\_Boxes} + Loss_{D^3}
\end{align}
where ${\bf{B}}_{ij}$ is an element located at $i$th row and $j$th column in ${\bf{B}}$. ${\bf{B}}_{ij}$ is equal to ${\bf{B}}_{ji}$ in a symmetric matrix so the output of D$^3$ should be divided by two.
When the model is equipped D$^3$, duplicate detection may be within limits. However, in sports video, the quality of MOT could go a step further while making use of some special rules of sports, which pedestrian video is not in the possession of.

\subsection{Rally-Hungarian Algorithm}

\begin{figure}
\centering
\includegraphics[width=10.55cm]{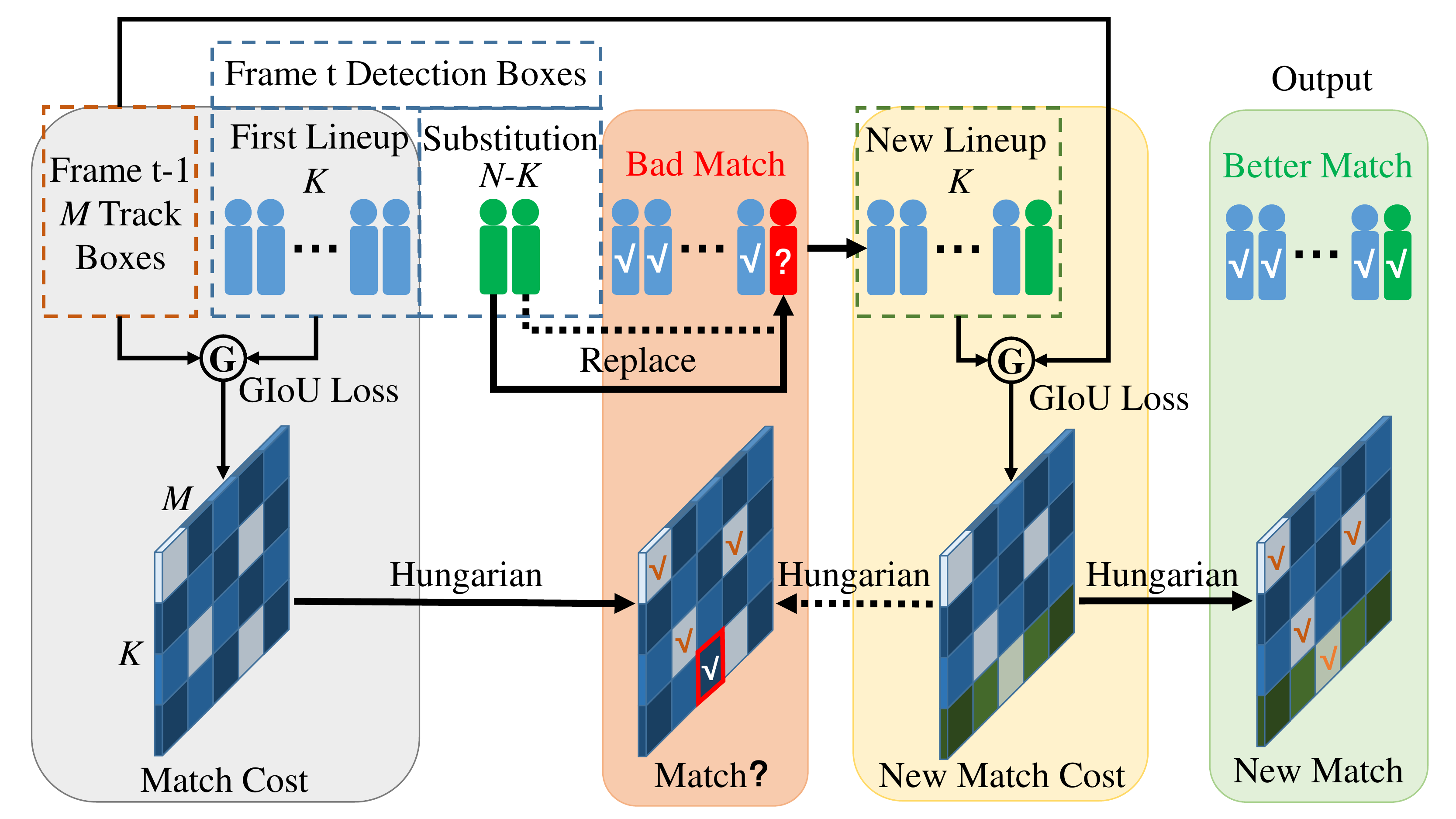}
\caption{Through a tracking model, detection boxes of frame $t$ and track boxes of frame $t-1$ are acquired. Then detection boxes are split into Lineup and Substitution. A rally of constructing match cost matrix by computing GIoU loss from Lineup and track boxes, getting  match pairs by applying Hungarian algorithm on the matrix, replacing the bad matching item with Substitution to form new Lineup is executed, which called Rally-Hungarian (RH). }
\label{fig:RH}
\end{figure}

Overview of Rally-Hungarian (RH) algorithm is shown in Figure \ref{fig:RH}. Through a tracking model, detection boxes of frame $t$ which are denoted as $B_{det}=\{{\bf{b}}_{i} |i=1,\cdots,N\}$, in which the elements are sorted in descending order by detection score, and track boxes of frame $t-1$ as $B_{track}=\{{\bf{b}}_{j} |j=1,\cdots,M\}$ are acquired, as the defination of boxes set applied in Section \ref{sec:DDD}. According to a rule of team sports like volleyball that the number of players on the court is fixed, we split detection boxes set $B_{det}$ to Lineup and Substitution. The top $K$ elements in $B_{det}$ are regarded as Lineup $B_{lineup}=\{{\bf{b}}_{i} |i=1,\cdots,K\}$ and the rest elements as Substitution $B_{sub}=\{{\bf{b}}_{k} |k=K+1,\cdots,N\}$. Then we construct match cost matrix ${\bf{C}}$ by computing GIoU loss from $B_{lineup}$ and $B_{track}$ as following:
\begin{align}
\label{C}
{\bf{C}} = 1 - GIoU(B_{lineup},B_{track})
\end{align}
where GIoU is the same as formula (\ref{giou}). Then utilizing Hungarian Algorithm on ${\bf{C}}$, we could acquire a set of match indices pairs $P=\{(i,j)|i\in[1,K];j\in[1,M]\}$ in which both $i$ and $j$ are monochronic as follows:
\begin{align}
\label{P}
P = Hungarian({\bf{C}})
\end{align}
$P$ is labeled by check marks. If ${\bf{b}}_{i}$ and ${\bf{b}}_{j}$ belong to one individual, ${\bf{C}}_{ij}$ should be a relatively small value as tint in Figure \ref{fig:RH}, which means an individual is tracked. However, if an outlier is chosen, as marked red, a row where the value is occupying should be replaced. 

Here, we explain how rows with outliers can be replaced. According to the Hungarian algorithm, this value is the best match between row $i$ and column $j$. It means that of all the mismatched columns in the row in which the above outlier is located, the value is the minimum. It is also not considered a new target for the current space, i.e. sports video. Two inferences can then be drawn. First, there must be a value smaller than the outlier that exists among all the matching columns in the row, and the smaller value does not match. That is to say, there is another better matching row in one column where the smaller value is located. Secondly, this outlier is exactly the minimum value of this row, indicating a poor quality of the match. As a consequence, a row, or a detection box, with an abnormal value could be replaced.

Then the bad matching detection box in $B_{lineup}$, regarded as $B_{bad}$, could be replaced by a substitution in $B_{sub}$, and a new lineup set $B_{new}$ is composed as follows:
\begin{align}
\label{B}
B_{new} = (B_{lineup}\setminus{B_{bad}})\cup B_{sub} = (B_{lineup}\setminus{{\bf{b}}_{i}})\cup{\bf{b}}_{k}
\end{align}
Looping formula (\ref{C}), (\ref{P}), (\ref{B}) as the dash arrows until each ${\bf{C}}_{ij}$ becomes acceptable or the $B_{sub}$ is empty. In the end, we get a better match pair set. In the field of volleyball, a rally means a round will not stop until the ball touch floor, like a loop. So we name our matching strategy as Rally-Hungarian (RH) algorithm, and ``R'' may have a dual meaning of ``Replace''.

\section{RallyTrack Dataset}
\label{sec:data}
In this section, RallyTrack Dataset will be introduced in detail.

\begin{figure}
\begin{tabular}{ccc}
\includegraphics[width=3.85cm]{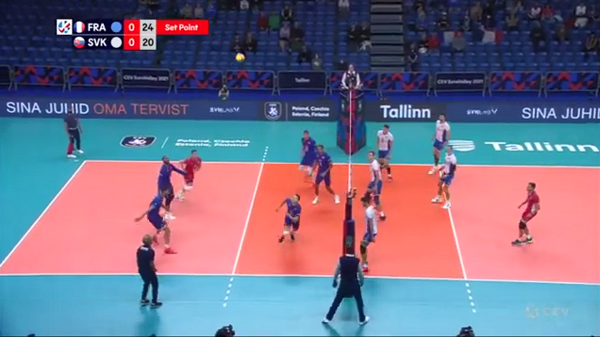}&
\includegraphics[width=4.03cm]{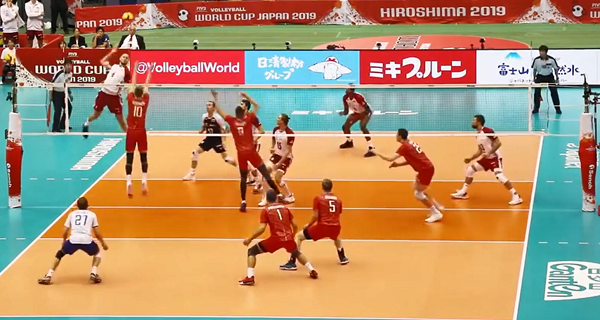}&
\includegraphics[width=3.85cm]{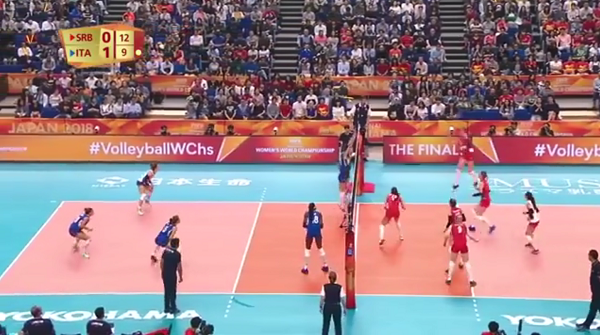}
\\(a)&(b)&(c)
\end{tabular}
\caption{Some samples in RallyTrack: (a) broadcast view of men's game; (b) fixed view of men's game; (c) broadcast view of women's game. }
\label{fig:RallyTrack}
\end{figure}

We annotate a RallyTrack dataset based primarily on sports videos for the MAT task as shown in Figure \ref{fig:RallyTrack}. The videos are from different views, broadcast or fixed, and different gender, men or women, of volleyball games. To guarantee training data and test data are not crossed, only will games from different Series be set as train and test. For example, if both games come from Rio 2016 Olympic Games, they should be put into a train set or test set together, even if each team is different. Some details of RallyTrack are then displayed in Table \ref{table:Datasets}. All of our data are labeled in MOT17 annotation format. As the test set's ground truths of MOT17 are not published, only the train set is calculated. In this table, column F/V refers to the number of frames showing more frames in RallyTrack than in each MOT17 video. Column O/F means objects per frame which show that individuals in RallyTrack are less than MOT17. Column T/V means tracks per video which show that trajectories in RallyTrack are also less than MOT17. However, given the overall situation of O/F and T/V, O/F is closer to T/V in RallyTrack than in MOT17, suggesting that RallyTrack has a longer personal trajectory than MOT17.

\begin{table}
\caption{Datasets comparison between MOT17 and RallyTrack. F/V means frames per video. O/F means objects per frame. T/V means tracks per video.}
\begin{center}
\begin{tabular}{ccccccccc}
\hline
\noalign{\smallskip}
Dataset & Subset &\, Videos\, &\, Frames \,&\, F/V \,& Objects  &\, O/F \,&  Tracks &\, T/V \, \\
\noalign{\smallskip}
\hline
\noalign{\smallskip}
MOT17 & \makecell[c]{Train\\Test\\Total}
& \makecell[c]{7\\7\\14}  
& \makecell[c]{5316\\5919\\11235}
& \makecell[c]{759.4\\845.6\\802.5}
& \makecell[c]{85828\\-\\-}
& \makecell[c]{16.1\\-\\-}
& \makecell[c]{546\\-\\-}
& \makecell[c]{78\\-\\-}\\
\noalign{\smallskip}
\hline
\noalign{\smallskip}
\,RallyTrack \,& \makecell[c]{Train\\Test\\Total} 
& \makecell[c]{10\\10\\20}  
& \makecell[c]{8104\\9757 \\17861}
& \makecell[c]{810.4\\975.7 \\893.1}
& \makecell[c]{68449\\91126\\159575}
& \makecell[c]{8.5\\9.3\\8.9}
& \makecell[c]{122\\126\\248}
& \makecell[c]{12.2\\12.6\\12.4}\\
\hline
\end{tabular}
\end{center}
\label{table:Datasets}
\end{table}

\section{Experimental Results}
\label{sec:results}

\subsection{Experimental Setup}

We evaluate D$^3$ on benchmarks: RallyTrack, MOT17, MOT16, MOT20, and DanceTrack. Following previous practice \cite{ZhouKK20,abs-2012-15460}, we split all the training sets of the MOT-series into two parts, one for training and the other for validation. The operation is samely applied on RallyTrack where half of the train set will be used and the whole test set will be tested. The widely-used MOT metrics set \cite{BernardinS08} is adopted for quantitative evaluation where multiple objects tracking accuracies (MOTA) is the primary metric to measure the overall performance. What is more, the higher order tracking accuracy (HOTA) \cite{luiten2020IJCV,luiten2020trackeval}, which explicitly balances the effect of performing accurate detection, association and localization into a single unified metric for comparing trackers, is also applied. While evaluating RH, only RallyTrack is used. 

For a fair comparison, we maintain most of the settings in TransTrack \cite{abs-2012-15460}, such as ResNet-50 \cite{HeZRS16} network backbone, Deformable DETR \cite{ZhuSLLWD21} based transformer structure, AdamW \cite{LoshchilovH19} optimizer, batch size 16. The initial learning rate is 2e-4 for the transformer and 2e-5 for the backbone. The weight decay is 1e-4. All transformer weights are initialized with Xavier-init \cite{GlorotB10}. The backbone model is pre-trained on ImageNet \cite{DengDSLL009} with frozen batch-norm layers \cite{IoffeS15}. Data augmentation includes random horizontal, random crop, scale augmentation, and resizing the input images whose shorter side is by 480-800 pixels while the longer side is by at most 1333 pixels. When the model is trained for 150 epochs, the learning rate drops by a factor of 10 at the 100th epoch.

\subsection{Experiments on RallyTrack and Others}

Our models are evaluated on RallyTrack as shown in Table \ref{table:RallyTrack}. In this table, the original TransTrack (TT) \cite{abs-2012-15460} model based on Deformable Transformer  \cite{ZhuSLLWD21} is regarded as a baseline, Rally-Hungarian (RH) algorithm and Duplicate Detection Decontaminator (D$^3$) could be evaluated respectively or jointly. In this table, TT with both D$^3$ and RH gets a stunning 9.2 rating on MOTA, 7.0 on IDF1 and 4.5 on HOTA to baseline. The results show that our methods are not only good at detecting multiple athletes but also associating them. It is mainly caused by decontaminating the duplicate detections and many athletes are correctly detected and tracked. 

\begin{table*}
\caption{Experiments on RallyTrack. Our method makes an amazing 9.2 promotion on MOTA, 7.0 on IDF1 and 4.5 on HOTA to baseline TransTrack (TT). Best in bold.}
\begin{center}
\begin{tabular}{lcccccccccc}
\hline
\noalign{\smallskip}
Model & MOTA$\uparrow$ & IDF1$\uparrow$ & MOTP$\uparrow$ & MT$\uparrow$ &   FP$\downarrow$ &  FN$\downarrow$ & IDS$\downarrow$ & HOTA$\uparrow$ & DetA$\uparrow$ & AssA$\uparrow$\\
\noalign{\smallskip}
\hline
\noalign{\smallskip}
TT \cite{abs-2012-15460}  & 59.5 & 28.8 & 77.8 & 70.6 & 15370 &  19489 & 2049 & 27.9 &  51.9 & 15.2 \\
TT+RH                     & 62.0 & 33.3 & 77.8 & 66.7 & 12557 &  20310 & \bf{1788} & 30.2  & 52.3 &  17.7 \\
TT+D$^3$    & 66.4 & 29.7 & 78.1 & \bf{78.6} & 13676 &  \bf{14848} & 2107 & 29.2 & 55.8 & 15.5 \\
TT+D$^3$+RH & \bf{68.7} & \bf{35.8} & \bf{78.1} & 77.0 & \bf{11359} & 15350 & 1847 & \bf{32.4} & \bf{56.3} & \bf{18.9}\\
\hline
\end{tabular}
\end{center}
\label{table:RallyTrack}
\end{table*}

\begin{table}[t]
\caption{Experiments on MOT17, MOT16, MOT20, and DanceTrack. Our method converges faster. Best in bold.}
\begin{center}
\begin{tabular}{cccccccccccc}
\hline
\noalign{\smallskip}
D$^3$ & Epoch & MOTA$\uparrow$ & IDF1$\uparrow$ & MOTP$\uparrow$ & MT$\uparrow$ &   FP$\downarrow$ &  FN$\downarrow$ & IDS$\downarrow$ & HOTA$\uparrow$ & DetA$\uparrow$ & AssA$\uparrow$\\
\noalign{\smallskip}
\hline
\noalign{\smallskip}
w/o    & 150 & 65.1 & 63.6 & 81.9 & 36.8 & 1918 &  16440 & \bf{438} & 52.6 & 54.0 & 51.7 \\
0.010 & 150 & 65.3 & 62.9 & 82.2 & 38.3 & 1849 &  \bf{16358} & 457 & 53.0 & 54.4 & 52.1\\
w/o    & 30 & 64.9 & 62.6 & 82.0 & 36.3 & 1862 &  16537 & 477 & 52.1 & 53.9 & 50.7\\
0.010 & \bf{30} & \bf{65.3} & \bf{63.6} & \bf{82.2} & \bf{38.6} & \bf{1833} &  16398 & 480 & \bf{53.4} & \bf{54.5} & \bf{52.8}\\
\noalign{\smallskip}
\hline
mot16\\
\hline
\noalign{\smallskip}
w/o    & 30 & 64.1 & 61.3 & 81.8 & 40.7 & \bf{2434} &  16186 & \bf{544} & 50.9 & 53.6 & 48.9 \\
0.011 & \bf{30} & \bf{65.3} & \bf{61.6} & 81.8 & 40.7 & 2578 &  \bf{15328} & 601 & \bf{52.3} & \bf{55.0} & \bf{50.3} \\
\noalign{\smallskip}
\hline
mot20\\
\hline
\noalign{\smallskip}
w/o    & 30 & 72.5 & 63.2 & 82.9 & 51.6 & 12882 &  153K & 2978 & 52.4 & 59.4 & 46.3 \\ 
0.017 & \bf{30} & \bf{73.2} & \bf{64.6} & 82.9 & \bf{53.4} & \bf{12831} &  \bf{149K} & \bf{2808} & \bf{53.6} & \bf{60.1} & \bf{47.9}\\
\noalign{\smallskip}
\hline
DT\\
\hline
\noalign{\smallskip}
w/o    & 50 & 76.5 & 39.4 & 85.2 & \bf{70.7} & 19087 &  29130 & 4795 & \bf{38.9} & \bf{66.8} & 22.9 \\
0.012 & 50 & 76.3 & 37.4 & 84.8 & 68.5 & 19432 &  \bf{28947} & 5026 & 37.1 & 66.4 & 21.0\\
w/o    & 25 & \bf{76.6} & 37.6 & \bf{85.2} & 70.0 & 18710 &  29348 & \bf{4685} & 38.1 & 67.1 & 21.8\\
0.012 & \bf{25} & 76.5 & \bf{39.4} & 84.9 & 67.8 & \bf{18518} &  29557 & 4808 & 38.7 & 66.2 & \bf{22.9}\\
\hline
\end{tabular}
\end{center}
\label{table:MOT17}
\end{table}

MOT17 is another dataset mainly used to measure the effectiveness of D$^3$ as shown in Table \ref{table:MOT17}. 
In this dataset, the main function of our approach is to reduce training time. The hyperparameters in the first column mean Lower Bound (LB) in D$^3$. LB is chosen according to different self-GIoU losses from different datasets. Different self-GIoU losses are caused by different resolutions of videos.
In this table, by actively eliminating duplicate detection, D$^3$ can save 80 percent of TT's training time, making the model converge faster from 150 down to 30 epochs. Instead, too many training epochs can lead to overfitting. We then demonstrated the priority of our approach by experimenting directly with MOT16 and MOT20 in the same setting as MOT17 for only 30 epochs.
DanceTrack (DT)  dataset is also measured with training on train set and testing on val set. As shown in Table \ref{table:MOT17}
only trained in 50 epochs could our method save 50 percent of the original training time and almost maintain the basic performance.

Datasets and solutions are massive for MOT after long-term development while it is not for MAT. We hope to provide a paradigm for MOT methods to easily extend to MAT. 
So D$^3$ is proposed as a connection between them. 
D$^3$ retains an almost complete structure of TT, allowing TT to expand for MAT (Table \ref{table:RallyTrack}) while maintaining the original MOT capabilities (Table \ref{table:MOT17}). 

\subsection{Ablation Study}
{\bf{Training Time and Lower Bound of D$^3$.}}
For an effective architecture, we tested numerous intuitive alternatives of D$^3$ from two main aspects, training epochs and lower bound (LB) settings, as shown in Table \ref{table:LB}. In this table, the results of the 40th and 150th training epoch, LB set as 0.010, 0.011, and 0.012 are displayed. It is clear that in row 5 and row 6 the performance achieves the best MOTA 66.4 and IDF1 29.7 when the training epoch is at 40th and the LB is determined at 0.011. The reason why MOTA is promoted is also obviously shown by FP and FN with a desirable gap, 6.9 percent at the 40th epoch, demonstrating that TT equipped with D$^3$ would make more correct predictions. Then another phenomenon, from row 2 to row 4, is that the result is very sensitive to LB. That a slightly 0.001 adjustment on LB could dorp 4.5 MOTA is warning the LB should be carefully set to decide whether duplicate detection exists. 

\begin{table}[t]
\caption{Training Time and Lower Bound. Best in bold.}
\begin{center}
\begin{tabular}{lcccccccccccc}
\hline
\noalign{\smallskip}
Model  &LB &Epoch & MOTA$\uparrow$ & IDF1$\uparrow$   & MT$\uparrow$ &  FP$\downarrow$ &  FN$\downarrow$ & IDS$\downarrow$ & HOTA$\uparrow$ & DetA$\uparrow$ & AssA$\uparrow$\\
\noalign{\smallskip}
\hline
\noalign{\smallskip}
TT      & -     & 150 & 59.4 & 28.7 & 69.8 & 15520 &  19426 & 2052 & 28.1 & 51.9 & 15.4\\
TT+D$^3$  & 0.010 & 150 & 61.9 & 29.2 & 73.0 & 15393 &  17166 & 2164 & 28.9 & 53.4 & 15.9 \\
TT+D$^3$  & 0.011 & 150 & 66.3 & 29.5 & 78.6 & 13806 &  {\bf{14750}} & 2109 & 29.2 & 55.8 & 15.5\\
TT+D$^3$  & 0.012 & 150 & 61.8 & 29.0 & 74.6 & 15954 &  16540 & 2309 & 28.8 & 53.0 & 16.0\\
TT      & -     & 40  & 59.5 & 28.8 & 70.6 & 15370 &  19489 & {\bf{2049}} & 27.9 &  51.9 & 15.2 \\
TT+D$^3$  & 0.011 & 40  & {\bf{66.4}} & {\bf{29.7}} & {\bf{78.6}} & {\bf{13676}} & 14848 & 2107 & {\bf{29.2}} & {\bf{55.8}} & {\bf{15.5}} \\
\hline
\end{tabular}
\end{center}
\label{table:LB}
\end{table}

{\bf{Age L, Top K and Replacement of RH.}}
We mainly report the effect of age L, top K, and replacement of RH as shown in Tabel \ref{table:LKR}. Age L means that if a tracking box is unmatched, it keeps as an ``inactive'' tracking box until it remains unmatched for L consecutive frames. Inactive tracking boxes can be matched to detection boxes and regain their ID. Following \cite{abs-2012-15460}, we firstly choose $L=32$ and then lengthen L to 80  causes in sport video individuals may disappear in an image but are always in the court. The data are collected based on volleyball videos so we choose top $K=12$.  Finally, whether replace with substitution is also evaluated and variant replacing strategies are discussed next.

\begin{table}[t]
\caption{Age L, Top K and Replacement of RH.}
\begin{center}
\begin{tabular}{lcccccc}
\hline
\noalign{\smallskip}
          &\,\, L  \,&\,\, K \, &\, Replace \,&\, MOTA \,&\, IDF1 \,&\, HOTA\\
\noalign{\smallskip}
\hline
\noalign{\smallskip}
TT+D$^3$    & 32 & -  & No      &66.4 &29.7 &29.2\\
TT+D$^3$    & 80 & -  & No      &66.3 &32.2 &30.3\\
TT+D$^3$+RH \,& 32 & 12 & No      &67.7 &30.9 &30.3\\
TT+D$^3$+RH & 80 & 12 & No      &67.7 &33.2 &31.0\\
TT+D$^3$+RH & 80 & 12 & Yes     &\bf{68.7} &\bf{35.8} &\bf{32.4}\\
\hline
\end{tabular}
\end{center}
\label{table:LKR}
\end{table}

{\bf{Replacing Strategy of RH.}}
We sweep over 5 different replacing strategies (RS) of the RH algorithm as shown in Table \ref{table:RS}. In this table, we assume that there are $p$ items in $B_{bad}$, also regarded as to be replaced items, and $q$ items in Substitution. Delete No. means the number of being removed items in $B_{bad}$ , and ``1st Bad'' means delete the first item in $B_{bad}$. Replace No. means the number of being replaced items in $B_{sub}$. As the elements in $B_{sub}$ are already sorted in descending order by detection score, the first one has the highest detection score in $B_{sub}$, which is marked as ``1st Score''. When a bad item is replaced by a high score substitution, it is also able to get a bad match. So all the items in $B_{sub}$ could be replaced to find a good match, and the first item composing a good match is marked as ``1st Good''. Then time complexity of each strategy is also analyzed. In this table, the total $q$ is set as 3, so the FPSes are close.

\begin{table}[t]
\caption{Replacing Strategies of RH.}
\begin{center}
\begin{tabular}{cccccccc}
\hline
\noalign{\smallskip}
&\, Delete No. \,&\, Replace No. \,& Complexity &MOTA &\, IDF1 \,& HOTA &\,FPS\, \\
\noalign{\smallskip}
\hline
\noalign{\smallskip}
RS1 & $p$       & 1 (1st Score)       & $O(p)$  &67.5 & 31.9 & 30.4 &6.41\\
RS2 & $p$       & 1 (1st Good)        & $O(pq)$ &67.5 & 31.8 &{30.2} &6.43\\
RS3 & $p$       & $min\{p,q\}$       &\, $O(p\cdot min\{p,q\})$ \,&67.6 & 32.2 &30.4 &6.33 \\
RS4 & 1 (1st Bad)& 1 (1st Score)       & $O(1)$ &68.4 & 35.5 &\bf{32.5} &\bf{6.52}\\
RS5 & 1 (1st Bad)& 1 (1st Good)       & $O(q)$ &\bf{68.7} &\bf{35.8} &32.4 &6.44\\
\hline
\end{tabular}
\end{center}
\label{table:RS}
\end{table}

\begin{table}[t]
\caption{Experiments on basketball and soccer.}
\begin{center}
\begin{tabular}{lcccccccccc}
\hline
\noalign{\smallskip}
Basketball & MOTA$\uparrow$ & IDF1$\uparrow$ & MOTP$\uparrow$ & MT$\uparrow$ & FP$\downarrow$ &  FN$\downarrow$ & IDS$\downarrow$ & HOTA$\uparrow$ & DetA$\uparrow$ & AssA$\uparrow$ \\
\noalign{\smallskip}
\hline
TT \cite{abs-2012-15460}  & 39.7 & 12.5 & 72.9 & 20.0  & 3995 &  3396 & 636 & 14.8 &  43.1 & 5.2\\
TT+RH                     & 45.1 & 14.4 & 73.0 & 20.0  & 3079 &  3680 & 593 & 15.9 &  44.6 & 5.7\\
TT+D$^3$    & 53.2 & 12.5 & 74.7 & 20.0  & 2450 &  \bf{3205} & 605 & 15.2 &  \bf{48.6} & 4.8\\
TT+D$^3$+RH & \bf{54.4} & \bf{16.8} &\bf{74.7} & 20.0  & \bf{2166} &  3420 & \bf{520} & \bf{17.4} &  48.0 & \bf{6.3}\\
\hline
Soccer 
\\
\hline
\noalign{\smallskip}
TT \cite{abs-2012-15460}  & \bf{60.1} & 21.2 & 79.3 & 33.3  & 1232 &  \bf{4205} & \bf{327} &23.2 &\bf{51.4} &10.5 \\
TT+RH                     & 59.2 & \bf{24.0} & \bf{79.4} & \bf{42.9}  & 1287 &  4260 & 354 &\bf{24.1} &50.9 &\bf{11.4}\\
TT+D$^3$    & 55.7 & 19.3 & 78.9 & 33.3 & \bf{1156} &  4868 & 381 &20.8 &48.2 &9.0\\
TT+D$^3$+RH & 57.9 & 21.4 & 78.8 & 28.6 & 1175 &  4907 & 395 &22.0 &47.9 &10.2 \\
\hline
\end{tabular}
\end{center}
\label{table:bs}
\end{table}

\subsection{Details of basketball and soccer videos}
Additionally, extending the RallyTrack dataset to other sports, we labeled 1484 frames of basketball and 1422 frames of soccer and tested them as shown in Table \ref{table:bs}. 13384 objects are in basketball and 14802 objects in soccer. Results indicate that our method might be directly applied to basketball videos rather than soccer videos. The scene in basketball is more similar to volleyball than that in soccer, and occlusion is not serious in soccer mainly because the background is easily distinguishable and larger, as visualized in Figure \ref{fig:visual}.

\subsection{Visualization}
\begin{figure}[t]
\centering
\begin{tabular}{cccc}
\includegraphics[width=2.85cm]{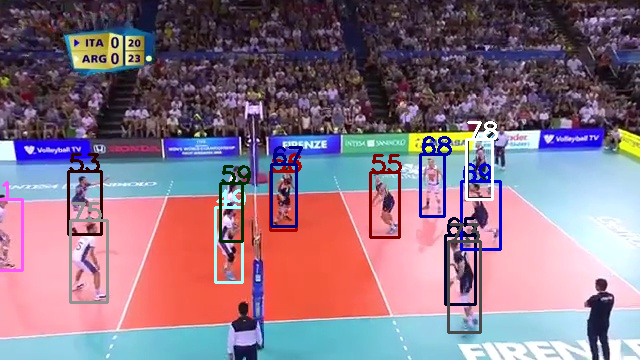}&
\includegraphics[width=2.85cm]{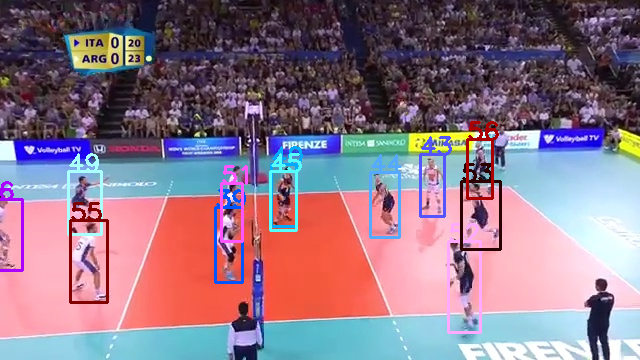}&
\includegraphics[width=2.85cm]{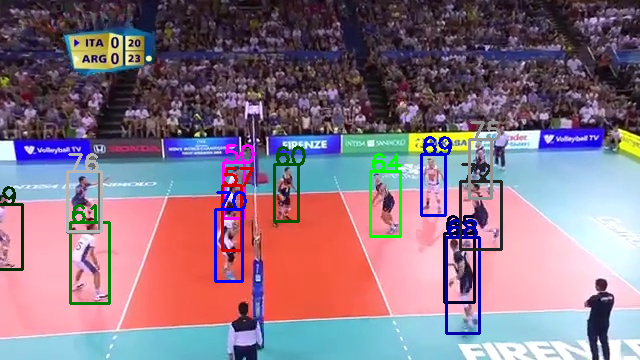}&
\includegraphics[width=2.85cm]{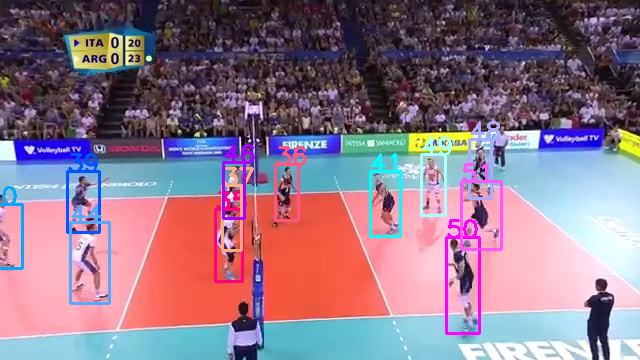}\\
(a) TT&(b) TT+RH&(c) TT+D$^3$&(d) TT+D$^3$+RH\\
\includegraphics[width=2.85cm]{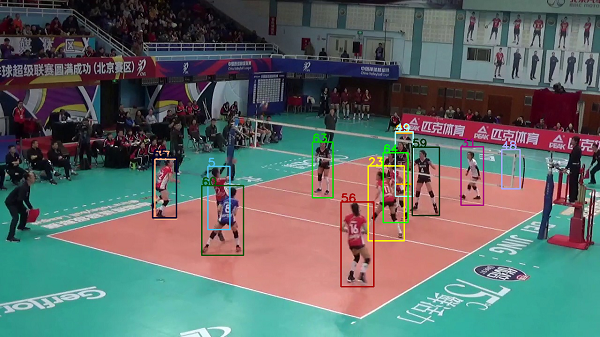}&
\includegraphics[width=2.85cm]{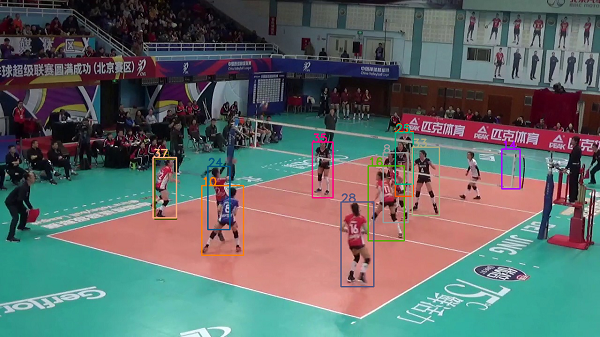}&
\includegraphics[width=2.85cm]{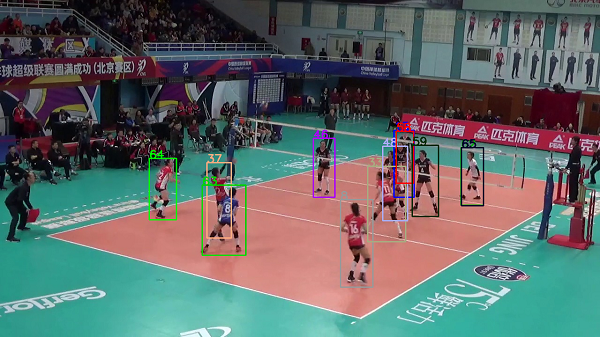}&
\includegraphics[width=2.85cm]{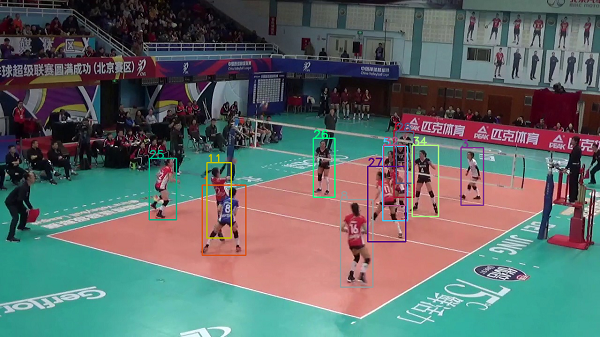}\\
(e) TT&(f) TT+RH&(g) TT+D$^3$&(h) TT+D$^3$+RH\\
\includegraphics[width=2.85cm]{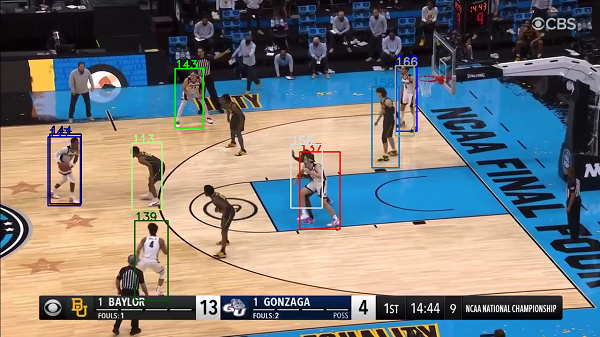}&
\includegraphics[width=2.85cm]{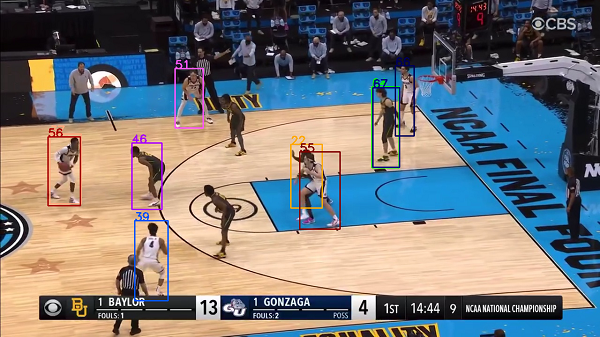}&
\includegraphics[width=2.85cm]{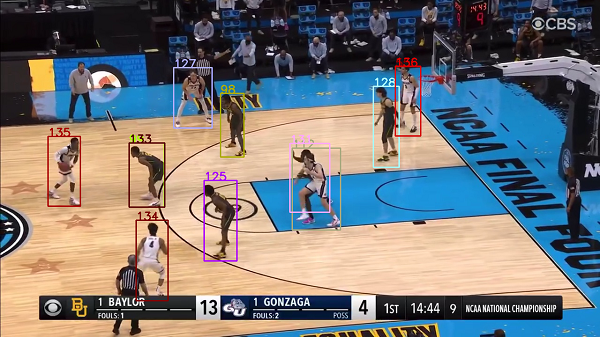}&
\includegraphics[width=2.85cm]{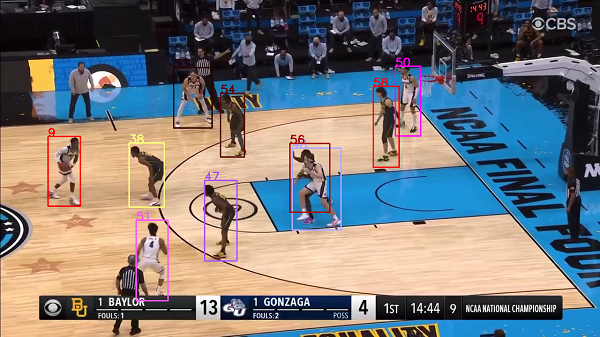}\\
(i) TT&(j) TT+RH&(k) TT+D$^3$&(l) TT+D$^3$+RH\\
\includegraphics[width=2.85cm]{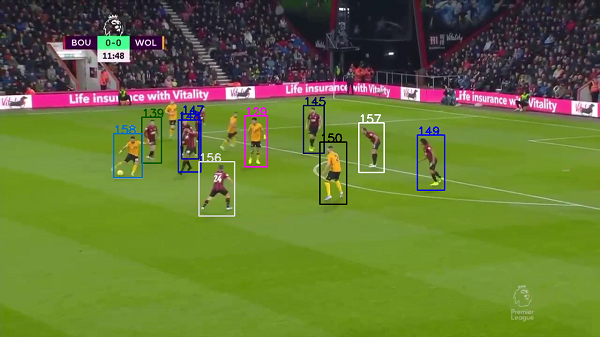}&
\includegraphics[width=2.85cm]{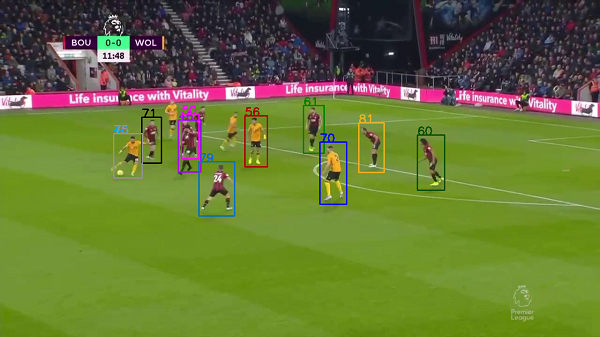}&
\includegraphics[width=2.85cm]{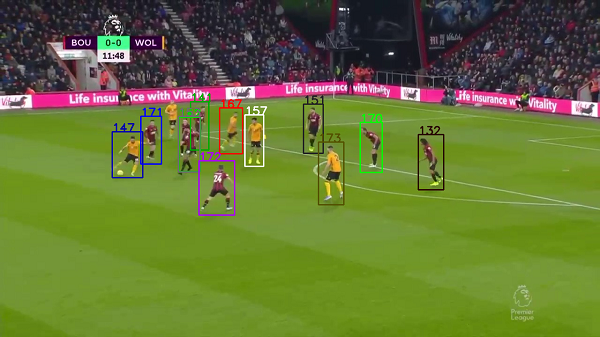}&
\includegraphics[width=2.85cm]{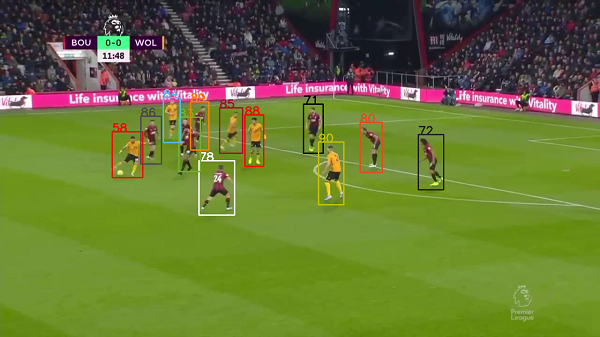}\\
(m) TT&(n) TT+RH&(o) TT+D$^3$&(p) TT+D$^3$+RH\\
\end{tabular}
\caption{Visualization of frame No.942 in test\_0100017 and frame No.715 in validation\_0160025  tracked by four different tracking models, and the models are directly applied to basketball and soccer videos.}
\label{fig:visual}
\end{figure}

We visualize two examples tracked by four different tracking models as shown in Figure \ref{fig:visual}. In \ref{fig:visual}(a), heavy duplicate detections happen and an object is missing while using a base model TT; then in \ref{fig:visual}(b), with RH, some duplicate detections disappear; moreover in \ref{fig:visual}(c), when equipped D$^3$, the missing object is found; finally in \ref{fig:visual}(d), combining D$^3$ and RH could get the best and the clearest tracking result. Then the best model is directly applied to basketball by setting $N=15$, $K=10$, $q=5$ in RH, and soccer by $N=20$, $K=15$, $q=5$ for the court of soccer is so large that usually not all individuals are visible. 

\section{Conclusion}
\label{sec:con}
In this paper, to address duplicate detection in MAT, we design a Duplicate Detection Decontaminator (D$^3$) which supervises the training procedure.
Then we design RH matching algorithm to go a step further on MAT. Experiments on our labeled RallyTrack show the priority of our methods. D$^3$ could also be utilized for saving training time on MOT17, MOT16, MOT20, and DanceTrack. Moreover, our model trained with volleyball data can be directly applied on other team sports videos like basketball or soccer, which may encourage more research exploring MAT applications.

\subsubsection{Acknowledgements} This work was supported by the National Natural Science Foundation of China under Grant U20B2069.
%
%
%
\bibliographystyle{splncs04}
\bibliography{egbib_hr}

\begin{thebibliography}{10}
\providecommand{\url}[1]{\texttt{#1}}
\providecommand{\urlprefix}{URL }
\providecommand{\doi}[1]{https://doi.org/#1}

\bibitem{BergmannML19}
Bergmann, P., Meinhardt, T., Leal{-}Taix{\'{e}}, L.: Tracking without bells and
  whistles. In: International Conference on Computer Vision. pp. 941--951
  (2019)

\bibitem{BernardinS08}
Bernardin, K., Stiefelhagen, R.: Evaluating multiple object tracking
  performance: The {CLEAR} {MOT} metrics. {EURASIP} J. Image Video Process.
  \textbf{2008} (2008)

\bibitem{BertinettoVHVT16}
Bertinetto, L., Valmadre, J., Henriques, J.F., Vedaldi, A., Torr, P.H.S.:
  Fully-convolutional siamese networks for object tracking. In: European
  Conference on Computer Vision. Lecture Notes in Computer Science, vol.~9914,
  pp. 850--865 (2016)

\bibitem{BewleyGORU16}
Bewley, A., Ge, Z., Ott, L., Ramos, F.T., Upcroft, B.: Simple online and
  realtime tracking. In: International Conference on Image Processing. pp.
  3464--3468 (2016)

\bibitem{CarionMSUKZ20}
Carion, N., Massa, F., Synnaeve, G., Usunier, N., Kirillov, A., Zagoruyko, S.:
  End-to-end object detection with transformers. In: European Conference on
  Computer Vision (2020)

\bibitem{ChavdarovaBBMJB18}
Chavdarova, T., Baqu{\'{e}}, P., Bouquet, S., Maksai, A., Jose, C.,
  Bagautdinov, T.M., Lettry, L., Fua, P., Gool, L.V., Fleuret, F.: {WILDTRACK:}
  {A} multi-camera {HD} dataset for dense unscripted pedestrian detection. In:
  Computer Vision and Pattern Recognition. pp. 5030--5039 (2018)

\bibitem{abs-2104-00194}
Chu, P., Wang, J., You, Q., Ling, H., Liu, Z.: Transmot: Spatial-temporal graph
  transformer for multiple object tracking. arxiv  \textbf{abs/2104.00194}
  (2021)

\bibitem{DaveKTSR20}
Dave, A., Khurana, T., Tokmakov, P., Schmid, C., Ramanan, D.: {TAO:} {A}
  large-scale benchmark for tracking any object. In: Vedaldi, A., Bischof, H.,
  Brox, T., Frahm, J. (eds.) European Conference on Computer Vision. Lecture
  Notes in Computer Science, vol. 12350, pp. 436--454 (2020)

\bibitem{abs-2003-09003}
Dendorfer, P., Rezatofighi, H., Milan, A., Shi, J., Cremers, D., Reid, I.D.,
  Roth, S., Schindler, K., Leal{-}Taix{\'{e}}, L.: {MOT20:} {A} benchmark for
  multi object tracking in crowded scenes. arxiv  \textbf{abs/2003.09003}
  (2020)

\bibitem{DengDSLL009}
Deng, J., Dong, W., Socher, R., Li, L., Li, K., Fei{-}Fei, L.: Imagenet: {A}
  large-scale hierarchical image database. In: Computer Vision and Pattern
  Recognition. pp. 248--255 (2009)

\bibitem{EllisF10}
Ellis, A., Ferryman, J.M.: {PETS2010} and {PETS2009} evaluation of results
  using individual ground truthed single views. In: International Conference on
  Advanced Video and Signal-based Surveillance. pp. 135--142 (2010)

\bibitem{GeigerLU12}
Geiger, A., Lenz, P., Urtasun, R.: Are we ready for autonomous driving? the
  {KITTI} vision benchmark suite. In: Computer Vision and Pattern Recognition.
  pp. 3354--3361 (2012)

\bibitem{GiancolaADG18}
Giancola, S., Amine, M., Dghaily, T., Ghanem, B.: Soccernet: {A} scalable
  dataset for action spotting in soccer videos. In: Computer Vision and Pattern
  Recognition Workshops. pp. 1711--1721 (2018)

\bibitem{GlorotB10}
Glorot, X., Bengio, Y.: Understanding the difficulty of training deep
  feedforward neural networks. In: International Conference on Artificial
  Intelligence and Statistics. {JMLR} Proceedings, vol.~9, pp. 249--256 (2010)

\bibitem{HeZRS16}
He, K., Zhang, X., Ren, S., Sun, J.: Deep residual learning for image
  recognition. In: Computer Vision and Pattern Recognition. pp. 770--778 (2016)

\bibitem{HoKPKK20}
Ho, K., Kardoost, A., Pfreundt, F., Keuper, J., Keuper, M.: A two-stage minimum
  cost multicut approach to self-supervised multiple person tracking. In: Asian
  Conference on Computer Vision. Lecture Notes in Computer Science, vol. 12623,
  pp. 539--557. Springer (2020)

\bibitem{IoffeS15}
Ioffe, S., Szegedy, C.: Batch normalization: Accelerating deep network training
  by reducing internal covariate shift. In: International Conference on Machine
  Learning. {JMLR} Workshop and Conference Proceedings, vol.~37, pp. 448--456
  (2015)

\bibitem{luiten2020trackeval}
Jonathon~Luiten, A.H.: Trackeval.
  \url{https://github.com/JonathonLuiten/TrackEval} (2020)

\bibitem{1960A}
Kalman, R.E.: A new approach to linear filtering and prediction problems.
  Journal of Basic Engineering  \textbf{82D},  35--45 (1960)

\bibitem{KongHW20}
Kong, L., Huang, D., Wang, Y.: Long-term action dependence-based hierarchical
  deep association for multi-athlete tracking in sports videos. {IEEE} Trans.
  Image Process.  \textbf{29},  7957--7969 (2020)

\bibitem{KongZRLH21}
Kong, L., Zhu, M., Ran, N., Liu, Q., He, R.: Online multiple athlete tracking
  with pose-based long-term temporal dependencies. Sensors  \textbf{21}(1),
  ~197 (2021)

\bibitem{Leal-TaixeMRRS15}
Leal{-}Taix{\'{e}}, L., Milan, A., Reid, I.D., Roth, S., Schindler, K.:
  Motchallenge 2015: Towards a benchmark for multi-target tracking. arxiv
  \textbf{abs/1504.01942} (2015)

\bibitem{LeeK020}
Lee, H., Kim, I., Kim, D.: {VAN:} versatile affinity network for end-to-end
  online multi-object tracking. In: Asian Conference on Computer Vision.
  Lecture Notes in Computer Science, vol. 12623, pp. 576--593. Springer (2020)

\bibitem{LinGGHD17}
Lin, T., Goyal, P., Girshick, R.B., He, K., Doll{\'{a}}r, P.: Focal loss for
  dense object detection. In: International Conference on Computer Vision. pp.
  2999--3007 (2017)

\bibitem{LoshchilovH19}
Loshchilov, I., Hutter, F.: Decoupled weight decay regularization. In:
  International Conference on Learning Representations (2019)

\bibitem{LuRVH20}
Lu, Z., Rathod, V., Votel, R., Huang, J.: Retinatrack: Online single stage
  joint detection and tracking. In: Computer Vision and Pattern Recognition.
  pp. 14656--14666 (2020)

\bibitem{luiten2020IJCV}
Luiten, J., Osep, A., Dendorfer, P., Torr, P., Geiger, A., Leal-Taix{\'e}, L.,
  Leibe, B.: Hota: A higher order metric for evaluating multi-object tracking.
  International Journal of Computer Vision pp. 1--31 (2020)

\bibitem{abs-2101-02702}
Meinhardt, T., Kirillov, A., Leal{-}Taix{\'{e}}, L., Feichtenhofer, C.:
  Trackformer: Multi-object tracking with transformers. arxiv
  \textbf{abs/2101.02702} (2021)

\bibitem{MilanL0RS16}
Milan, A., Leal{-}Taix{\'{e}}, L., Reid, I.D., Roth, S., Schindler, K.:
  {MOT16:} {A} benchmark for multi-object tracking. arxiv
  \textbf{abs/1603.00831} (2016)

\bibitem{NiuGT12}
Niu, Z., Gao, X., Tian, Q.: Tactic analysis based on real-world ball trajectory
  in soccer video. Pattern Recognit.  \textbf{45}(5),  1937--1947 (2012)

\bibitem{PangQLCLDY21}
Pang, J., Qiu, L., Li, X., Chen, H., Li, Q., Darrell, T., Yu, F.: Quasi-dense
  similarity learning for multiple object tracking. In: Computer Vision and
  Pattern Recognition. pp. 164--173 (2021)

\bibitem{PengWWWWTWLHF20}
Peng, J., Wang, C., Wan, F., Wu, Y., Wang, Y., Tai, Y., Wang, C., Li, J.,
  Huang, F., Fu, Y.: Chained-tracker: Chaining paired attentive regression
  results for end-to-end joint multiple-object detection and tracking. In:
  European Conference on Computer Vision. Lecture Notes in Computer Science,
  vol. 12349, pp. 145--161 (2020)

\bibitem{abs-1804-02767}
Redmon, J., Farhadi, A.: Yolov3: An incremental improvement. arxiv
  \textbf{abs/1804.02767} (2018)

\bibitem{RenHGS15}
Ren, S., He, K., Girshick, R.B., Sun, J.: Faster {R-CNN:} towards real-time
  object detection with region proposal networks. In: Conference and Workshop
  on Neural Information Processing Systems. pp. 91--99 (2015)

\bibitem{RezatofighiTGS019}
Rezatofighi, H., Tsoi, N., Gwak, J., Sadeghian, A., Reid, I.D., Savarese, S.:
  Generalized intersection over union: {A} metric and a loss for bounding box
  regression. In: Computer Vision and Pattern Recognition. pp. 658--666 (2019)

\bibitem{SunKDCPTGZCCVHN20}
Sun, P., Kretzschmar, H., Dotiwalla, X., Chouard, A., Patnaik, V., Tsui, P.,
  Guo, J., Zhou, Y., Chai, Y., Caine, B., Vasudevan, V., Han, W., Ngiam, J.,
  Zhao, H., Timofeev, A., Ettinger, S., Krivokon, M., Gao, A., Joshi, A.,
  Zhang, Y., Shlens, J., Chen, Z., Anguelov, D.: Scalability in perception for
  autonomous driving: Waymo open dataset. In: Computer Vision and Pattern
  Recognition. pp. 2443--2451 (2020)

\bibitem{abs-2111-14690}
Sun, P., Cao, J., Jiang, Y., Yuan, Z., Bai, S., Kitani, K., Luo, P.:
  Dancetrack: Multi-object tracking in uniform appearance and diverse motion.
  arxiv  \textbf{abs/2111.14690} (2021)

\bibitem{abs-2012-15460}
Sun, P., Jiang, Y., Zhang, R., Xie, E., Cao, J., Hu, X., Kong, T., Yuan, Z.,
  Wang, C., Luo, P.: Transtrack: Multiple-object tracking with transformer.
  arxiv  \textbf{abs/2012.15460} (2020)

\bibitem{TangWWLZW20}
Tang, P., Wang, C., Wang, X., Liu, W., Zeng, W., Wang, J.: Object detection in
  videos by high quality object linking. {IEEE} Trans. Pattern Anal. Mach.
  Intell.  \textbf{42}(5),  1272--1278 (2020)

\bibitem{VaswaniSPUJGKP17}
Vaswani, A., Shazeer, N., Parmar, N., Uszkoreit, J., Jones, L., Gomez, A.N.,
  Kaiser, L., Polosukhin, I.: Attention is all you need. In: Conference and
  Workshop on Neural Information Processing Systems. pp. 5998--6008 (2017)

\bibitem{VoigtlaenderKOL19}
Voigtlaender, P., Krause, M., Osep, A., Luiten, J., Sekar, B.B.G., Geiger, A.,
  Leibe, B.: {MOTS:} multi-object tracking and segmentation. In: Computer
  Vision and Pattern Recognition. pp. 7942--7951 (2019)

\bibitem{WangZLLW20}
Wang, Z., Zheng, L., Liu, Y., Li, Y., Wang, S.: Towards real-time multi-object
  tracking. In: Vedaldi, A., Bischof, H., Brox, T., Frahm, J. (eds.) European
  Conference on Computer Vision. Lecture Notes in Computer Science, vol. 12356,
  pp. 107--122 (2020)

\bibitem{WojkeBP17}
Wojke, N., Bewley, A., Paulus, D.: Simple online and realtime tracking with a
  deep association metric. In: International Conference on Image Processing.
  pp. 3645--3649 (2017)

\bibitem{Xu0ZH19}
Xu, J., Cao, Y., Zhang, Z., Hu, H.: Spatial-temporal relation networks for
  multi-object tracking. In: International Conference on Computer Vision. pp.
  3987--3997 (2019)

\bibitem{abs-1809-03327}
Xu, N., Yang, L., Fan, Y., Yue, D., Liang, Y., Yang, J., Huang, T.S.:
  Youtube-vos: {A} large-scale video object segmentation benchmark. arxiv
  \textbf{abs/1809.03327} (2018)

\bibitem{1955The}
Yaw, H.: The hungarian method for the assignment problem. In: Naval Res Logist
  Quart (1955)

\bibitem{YuLLLSY16}
Yu, F., Li, W., Li, Q., Liu, Y., Shi, X., Yan, J.: {POI:} multiple object
  tracking with high performance detection and appearance feature. In: European
  Conference on Computer Vision. Lecture Notes in Computer Science, vol.~9914,
  pp. 36--42 (2016)

\bibitem{YuCWXCLMD20}
Yu, F., Chen, H., Wang, X., Xian, W., Chen, Y., Liu, F., Madhavan, V., Darrell,
  T.: {BDD100K:} {A} diverse driving dataset for heterogeneous multitask
  learning. In: Computer Vision and Pattern Recognition. pp. 2633--2642 (2020)

\bibitem{abs-2105-03247}
Zeng, F., Dong, B., Wang, T., Chen, C., Zhang, X., Wei, Y.: {MOTR:} end-to-end
  multiple-object tracking with transformer. arxiv  \textbf{abs/2105.03247}
  (2021)

\bibitem{abs-2110-06864}
Zhang, Y., Sun, P., Jiang, Y., Yu, D., Yuan, Z., Luo, P., Liu, W., Wang, X.:
  Bytetrack: Multi-object tracking by associating every detection box. arxiv
  \textbf{abs/2110.06864} (2021)

\bibitem{ZhangWWZL21}
Zhang, Y., Wang, C., Wang, X., Zeng, W., Liu, W.: Fairmot: On the fairness of
  detection and re-identification in multiple object tracking. Int. J. Comput.
  Vis.  \textbf{129}(11),  3069--3087 (2021)

\bibitem{abs-1811-11167}
Zhang, Z., Cheng, D., Zhu, X., Lin, S., Dai, J.: Integrated object detection
  and tracking with tracklet-conditioned detection. arxiv
  \textbf{abs/1811.11167} (2018)

\bibitem{ZhouKK20}
Zhou, X., Koltun, V., Kr{\"{a}}henb{\"{u}}hl, P.: Tracking objects as points.
  In: European Conference on Computer Vision. Lecture Notes in Computer
  Science, vol. 12349, pp. 474--490 (2020)

\bibitem{abs-1904-07850}
Zhou, X., Wang, D., Kr{\"{a}}henb{\"{u}}hl, P.: Objects as points. arxiv
  \textbf{abs/1904.07850} (2019)

\bibitem{ZhuSLLWD21}
Zhu, X., Su, W., Lu, L., Li, B., Wang, X., Dai, J.: Deformable {DETR:}
  deformable transformers for end-to-end object detection. In: International
  Conference on Learning Representations (2021)

\end{thebibliography}
\end{document}